\documentclass[11pt,a4paper]{article}
\usepackage[utf8]{inputenc}
\usepackage[hyperref]{naaclhlt2019}
\usepackage{times}
\usepackage{latexsym}
\usepackage{tikz}
\usetikzlibrary{positioning}
\usepackage{subfigure}
\usetikzlibrary{calc}
\def\layersep{2cm}
\usepackage{url}
\usetikzlibrary{chains}
\usetikzlibrary{arrows,positioning} 

\tikzset{
    >=stealth',
    punkt/.style={
           rectangle,
           rounded corners,
           draw=black, very thick,
           text width=4.5em,
           minimum height=2em,
           text centered},
      long/.style={
           rectangle,
           rounded corners,
           draw=black, very thick,
           text width= 15em,
           minimum height=2em,
           text centered},
    pil/.style={
           ->,
           thick,
           shorten <=2pt,
           shorten >=2pt,}
}

\aclfinalcopy 

\title{Train, Sort, Explain: Learning to Diagnose Translation Models}

\author{\textbf{Robert Schwarzenberg$^\mathbf{1}$, David Harbecke$^\mathbf{1}$, Vivien Macketanz$^\mathbf{1}$}, \\ \textbf{Eleftherios Avramidis$^\mathbf{1}$, Sebastian Möller$^\mathbf{1,2}$} \\ $^1$German Research Center for Artificial Intelligence (DFKI), Berlin, Germany \\
$^2$Technische Universität Berlin, Berlin, Germany \\
{\tt \{firstname.lastname\}@dfki.de}
}

\date{}

\begin{document}
\maketitle

\begin{abstract}
Evaluating translation models is a trade-off between effort and detail. On the one end of the spectrum there are automatic count-based methods such as BLEU, on the other end linguistic evaluations by humans, which arguably are more informative but also require a disproportionately high effort. To narrow the spectrum, we propose a general approach on how to automatically expose systematic differences between human and machine translations to human experts. Inspired by adversarial settings,  we train a neural text classifier to distinguish human from machine translations. A classifier that performs and generalizes well after training should recognize systematic differences between the two classes, which we uncover with neural explainability methods. Our proof-of-concept implementation, DiaMaT, is open source. Applied to a dataset translated by a state-of-the-art neural Transformer model, DiaMaT achieves a classification accuracy of 75\% and exposes meaningful differences between humans and the Transformer, amidst the current discussion about human parity. 
\end{abstract}

\section{Introduction}
A multi-dimensional diagnostic evaluation of performance or quality often turns out to be more helpful for system improvement than just considering a one-dimensional utilitarian metric, such as BLEU \cite{papineni2002bleu}. This is exemplified by, for instance, the pioneering work of \citet{bahdanau2014neural}. The authors introduced the attention mechanism responding to the findings of \citet{cho2014properties} who reported that neural translation quality degraded with sentence length. The attention mechanism was later picked up by \citet{vaswani2017attention} for their attention-only Transformer model, which still is state of the art in machine translation (MT) \cite{bojar-EtAl:2018:WMT1}. 
Furthermore, while MT output approaches human translation quality and the claims for "human parity" \cite{Wu2016b,DBLP:journals/corr/abs-1803-05567} increase, multi-dimensional diagnostic evaluations can be useful to spot the thin line between the machine and the human.

Diagnostic (linguistic) evaluations require human-expert feedback, which, however, is very time-consuming to collect. For this reason, there is a need for tools that mitigate the effort, such as the ones developed by \citet{6061334,popovic11:hjerson,Berka2012,klejch2015mt}. 

In this paper we propose a novel approach for developing evaluation tools. Contrary to the above tools that employ string comparison methods such as BLEU, implementations of the new approach derive annotations based on a neural model of explainability. This allows both capturing of semantics as well as focusing on the particular tendencies of MT errors. Using neural methods for the evaluation and juxtaposition of translations has already been done by \citet{VisualizingNeuralMachineTranslationAttentionandConfidence}. Their method, however, can only be applied to attention-based models and their translations. In contrast, our approach generalizes to arbitrary machine and even human translations. After first discussing the abstract approach in the next section, we present a concrete open-source implementation, ``DiaMaT'' (from \textit{Dia}gnose \textit{Ma}chine \textit{T}ranslations). 

\section{Approach}
The proposed approach consists of the three steps (1) \textit{train}, (2) \textit{sort}, and (3) \textit{explain}.

\subsection{Step 1: Train}
In a first step, inspired by generative adversarial networks \cite{goodfellow2014generative,DBLP:journals/corr/WuXZTQLL17,DBLP:journals/corr/YangCWX17} we propose to train a model to distinguish machine from human translations. The premise is that if the classifier generalizes well after training, it has learned to recognize systematic or frequent differences  between the two classes (herinafter also referred to as ``class evidence''). Class evidence may be, for instance, style differences, overused n-grams but also errors. The text classifier can be implemented through various architectures, ranging from deep CNNs \cite{verydeepcun} to recurrent classifiers built on top of pre-trained language models \cite{howard2018universal}.

\subsection{Step 2: Sort}
In a second step, we suggest letting the trained classifier predict the labels of a test set which contains human and machine translations and then sort them by classification confidence. This is based on the assumption that if the classifier is very certain that a given translation was produced by a machine (translation moved to the top of the list in this step), then the translation should contain strong evidence for a class, i.e.~errors typical for only the machine. Furthermore, even if we are dealing with a very human-like MT output, which means that our classifier may only slightly perform above chance, sorting by classification confidence should still move the few systematic differences that the classifier identified to the top.

\begin{figure}[!htb]
\centering
\begin{tikzpicture}[shorten >=1pt,->,draw=black, node distance=\layersep, scale=.9, transform shape]
    \tikzstyle{every pin edge}=[<-,shorten <=1pt]
    \tikzstyle{neuron}=[circle,draw,minimum size=17pt,inner sep=0pt]
    \tikzstyle{annot} = [text width=7em]
        
    \node[neuron,fill=red!60!white] (I-1) at (1,0) {};
    \node[neuron,fill=blue!40!white] (I-2) at (2,0) {};
    \node[neuron,fill=red!10!white] (I-3) at (3,0) {};
    \node[neuron,fill=red!50!white] (I-4) at (4,0) {};

    \node[neuron,fill=red!20!white] (H-1) at (1.5 cm,\layersep) {};
    \node[neuron,fill=blue!30!white] (H-2) at (2.5 cm,\layersep) {};
    \node[neuron,fill=red!60!white] (H-3) at (3.5 cm,\layersep) {};

    \node[neuron,fill=red!50!white] (O) at (2.5 cm,\layersep*2) {};

    \draw (H-1) edge node[sloped, anchor=south, pos=0.5, auto=false, above=-2pt] {} (I-1);
	\foreach \source in {2,3}
        \draw (H-\source) edge (I-\source);
    \foreach \source/\dest in {1/2,2/3,3/4}
        \draw (H-\source) edge (I-\dest);

    \draw (O) edge node[sloped, anchor=south, pos=0.5, auto=false, above=-2pt] {} (H-1);
    \foreach \dest in {2,3}
        \draw (O) edge (H-\dest);

    \node[annot,right of=O, node distance=3.5cm] (o) {positive output};
    \node[annot,right of=H-3, node distance=2.5cm] (hl) {hidden layer contributions};
    \node[annot,right of=I-4] (wvc) {word vector \\ contributions};
    
\end{tikzpicture}
\caption{Contributions propagated from output to input space. Colors represent positive (red) and negative (blue) contributions. The Figure is adapted from \citet{kindermans2018learning}.}
\label{fig:backprobcontributions}
\end{figure}
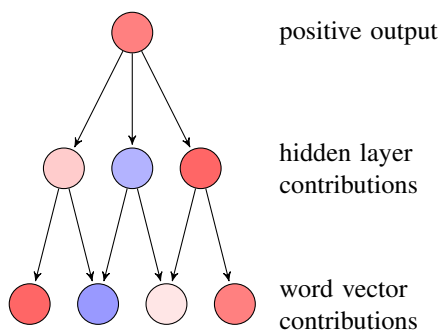
\subsection{Step 3: Explain}
\citet{arras2016explaining, arras-plos17, ArrWASSA17} demonstrated the data exploratory power of explainability methods in Natural Language Processing (NLP). This is why in a third step, we propose to apply an explainability method to uncover and visualize the class evidence on which the classifier based its decisions. Our definition of an explanation follows \citet{MonDSP18}, who define it as ``the collection of features of the interpretable domain, that have contributed for a given example to produce a decision (e.g.~classification or regression).''\footnote{\citet{MonDSP18} distinguish between explainability and interpretability. Interpretability methods also hold potential for the approach. For brevity, we limit ourselves to explainability methods here.} In our case the interpretable domain is the plain text space. There exist several candidate explainability methods, one of which we present in the following as an example. 

\begin{figure}[!htb]
    \centering
    \includegraphics[scale=.5]{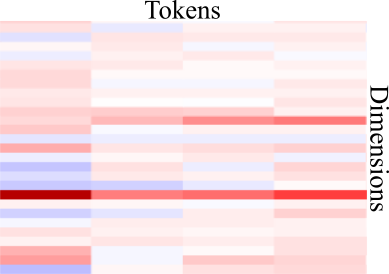}
    \caption{A heatmap of contribution scores in word vector space over a sequence of tokens. Red means positive contribution (score $>$ 0), blue means negative contribution (score $<$ 0).}
    \label{fig:heatmap}
\end{figure}

\begin{figure*}[!htb]
    \centering
    \includegraphics[width=.8\textwidth]{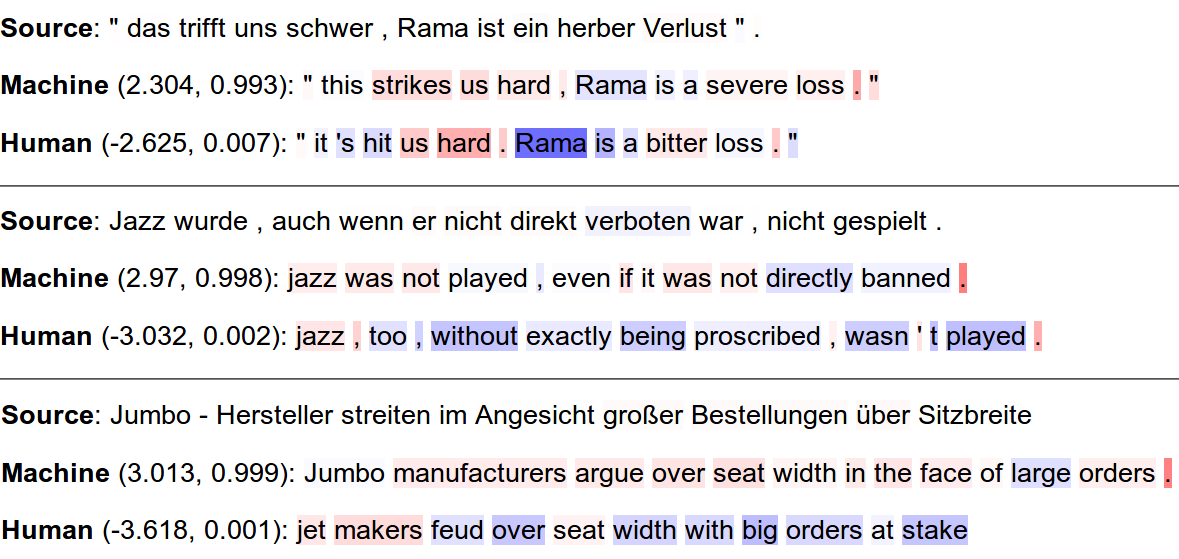}
    \caption{Screenshot of demonstrative results in DiaMaT. Filters that allow the user to analyse the corpus are not shown. The bold label is the true label. The activations of the machine neuron are shown in brackets; on the left the unnormalized logit activation, on the right the softmax activation. Positive logits and softmax probabilities greater 0.5 indicate machine evidence, as do tokens highlighted in red. Consequently, blue indicates evidence for the human. The more intense the colour, the stronger the evidence.}
    \label{fig:screenshot}
\end{figure*}

\subsubsection{Explainability and Interpretability Methods for Data Exploration}
\label{sec:expldataexpl}
In their tutorial paper, \citet{MonDSP18} discuss several groups of explainability methods. One group, for instance, identifies how sensitively a model reacts to a change in the input, others extract patterns typical for a certain class.   
Here, we discuss methods that propagate back contributions.

The contribution flow is illustrated in Fig.~\ref{fig:backprobcontributions}. At the top, the depicted binary classifier produced a positive output (input classified as class one). The classification decision is based on the fact that in the previous layer, the evidence for class one exceeded the evidence for class zero: The left and the right neuron contributed positively to the decision (reddish), whereas the middle neuron contributed negatively (blueish). 
Several explainability methods, such as Layerwise Relevance Propagation \cite{bach-plos15} or PatternAttribution \cite{kindermans2018learning}, backtrack contributions layer-wise. The methods have to preserve coherence over highly non-linear activation functions. 
Eventually, contributions are projected into the input space where they reveal what the model considers emblematic for a class. This is what we exploit in step 3.

Explainability methods in NLP \cite{arras2016explaining,  arras-plos17, ArrWASSA17,  harbecke2018learning} are typically used to first project scores into word-vector space resulting in heat maps as shown in Fig.~\ref{fig:heatmap}. To interpret them in plain text space, the scores are summed over the word vector dimensions to compute RGB values for each token, resulting in plain text heatmaps as shown in Fig.~\ref{fig:screenshot}.

\section{Implementation}
\begin{figure}[!htb]
    \centering
\begin{tikzpicture}[node distance=1cm, auto, scale=.9, transform shape]
 \node[punkt] (human) {Translation};
 \node[punkt, right=0.5cm of human] (machine) {Source};
 \node[punkt, right=0.5cm of machine] (source) {Translation};
 \node[punkt, above=0.5cm of human] (conv_human) {Conv};
 \node[punkt, above=0.5cm of machine] (conv_machine) {Conv};
 \node[punkt, above=0.5cm of source] (conv_source) {Conv};
 \node[long, above=0.5 of conv_machine](maxpoolconc) {Max Pooling \& Concatenation};
 \node[punkt, above=0.5 of maxpoolconc] (fc){FC$^{2 \times F}$};
 \draw[->] (human) -- (conv_human);
 \draw[->] (machine) -- (conv_machine); 
 \draw[->] (source) -- (conv_source);
  \draw[->] (conv_human) -- (maxpoolconc);
 \draw[->] (conv_machine) -- (maxpoolconc); 
 \draw[->] (conv_source) -- (maxpoolconc);
 \draw[->] (maxpoolconc) -- (fc);
\end{tikzpicture}
    \caption{Architecture of the text classifier.} 
    \label{fig:model_architecture}
\end{figure}
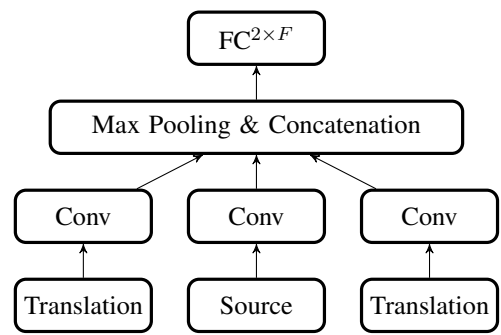

For step 1 (training phase), DiaMaT\footnote{Source code, data and experiments are available at \url{https://github.com/dfki-nlp/diamat}.} deploys a CNN text classifier, the architecture of which is depicted in Fig.~\ref{fig:model_architecture}.  The classifier consumes three embeddings: the embedding of a source and two translations of the source, one by a machine and one by a human. It then separately convolves over the embeddings and subsequently applies max pooling to the filter activations. The concatenated max features are then passed to the last layer, a fully connected layer with two output neurons. The left neuron fires if the machine translation was passed to the left input layer, the right neuron fires if the machine translation was passed to the right input layer. Note that this layer allows the model to combine features from all three inputs for its classification decision.

For step 2 (sorting phase), DiaMaT offers to sort by unnormalized logit activations or by softmax activations. Furthermore, one can choose to use the machine neuron activation or the human neuron activation as the sorting key. 

For step 3 (explaining phase), DiaMaT employs the iNNvestigate toolbox  \cite{alber2018innvestigate} in the back-end that offers more than ten explainability methods: Replacing one method with another only requires to change one configuration value in DiaMaT, before repeating step 3 again. In step 3, DiaMaT produces explanations in the form of $(token,\ score)$ tuple lists that are consumed by a front-end server which visualizes the scores as class evidence (see Fig.~\ref{fig:screenshot}).\footnote{The front-end was inspired by the demo LRP server of the Fraunhofer HHI insitute \url{https://lrpserver.hhi.fraunhofer.de/text-classification}, last accessed 2019-01-31.}

\section{Datasets and Experiments}
\label{sec:experiments}
We tested DiaMaT on a corpus translated by an NMT Transformer engine \cite{vaswani2017attention} conforming to the WMT14 data setup \cite{bojar-EtAl:2014:W14-33}. 
The NMT model was optimized on the test-set of WMT13 and an ensemble of 5 best models was used. 
It was trained using OpenNMT \cite{klein2017opennmt}, including Byte Pair Encoding \cite{sennrich_2015_neural} but no back-translation, achieving 32.68 BLEU on the test-set of WMT14.

Next, we trained the CNN text classifier sketched in Fig.~\ref{fig:model_architecture} for which we randomly drew 1M training samples (human references and machine translations alongside their sources) from the WMT18 training data \cite{bojar-EtAl:2018:WMT1}, excluding the WMT14 training data. The validation set consisted of 100k randomly drawn samples from the same set and we drew another 100k samples randomly for training the explainability method of choice, PatternAttribution, which learns explanations from data.  All texts were embedded using pre-trained fastText word vectors \citep{grave2018learning}.

We evaluated the classifier on around 20k samples drawn from the official test sets, excluding WMT13. On this test set, the classifier achieved an accuracy of 75\%, which is remarkable, considering the ongoing discussion about human parity \cite{Wu2016b,DBLP:journals/corr/abs-1803-05567}. We also used this test set for steps 2 and 3. Thus, neither the translation model, nor the text classifier, nor the explainability method encountered this split during training. For step 2, the machine translation was always passed to the right input layer and contributions to the right output neuron were retrieved with PatternAttribution.\footnote{In order to visualize evidence for the human (blue), positive contributions in the left input needed to be inverted.} We then sorted the inputs by the softmax activation of the machine neuron, which moved inputs for which the classifier is certain that it has identified the machine correctly to the top.

\section{Demonstration and Observations}
We observed that the top inputs frequently contained sentences in which DiaMaT considered the token after a sentence-ending full stop strong evidence for the human (Fig.~\ref{fig:screenshot}, top segment). We take this as evidence that DiaMaT correctly recognized that the human generated multiple sentences instead of a single one more often than the machine did. At this point, we cannot, however, offer an explanation for why the token preceding the punctuation mark is frequently considered evidence for the machine. 

Furthermore, DiaMaT also regarded reduced negations (``n't'') as evidence for the human (see Fig.~\ref{fig:screenshot}, middle segment) which again is reflected in the statistics. The machine tends to use the unreduced negation more frequently. 

The last segment in Fig.~\ref{fig:screenshot} shows how DiaMaT points to the fact that the machine more often produced sentence end markers than the human in cases where the source contained no end marker. The claims above are all statistically significant in the test set, according to a $\chi^{2}$ test with $\alpha = 0.001$.

\section{Future Work}
The inputs in Fig.~\ref{fig:screenshot} contain easily readable evidence. There is, however, also much evidence that is hard to read. In general, we can assume that with increasing architectural complexity, more complex class evidence can be uncovered, which may come at the cost of harder readability.

In the future, it is worth exploring how different architectures and model choices affect the quality, complexity and readability of the uncovered evidence. For instance, one direction would be to to train the classifier on top of a pretrained language model \citep{howard2018universal, devlin_bert} which could improve the classification performance. Furthermore, other explainability methods should also be tested.

\section{Conclusion}
We presented a new approach to analyse and juxtapose translations. Furthermore, we also presented an implementation of the approach, DiaMaT. DiaMaT exploits the generalization power of neural networks to learn systematic differences between human and machine translations and then takes advantage of neural explainability methods to uncover these. It learns from corpora containing millions of translations but offers explanations on sentence level. In a stress test, DiaMaT was capable of exposing systematic differences between a state-of-the-art translation model output and human translations.

\section*{Acknowledgements}
This research was partially supported by the German Federal Ministry of Education and Research through the project DEEPLEE (01IW17001). We would also like to thank the anonymous reviewers and Leonhard Hennig for their helpful feedback.

\bibliography{naaclhlt2019}
\bibliographystyle{acl_natbib}

\appendix
\label{sec:appendix}
\end{document}